\pdfoutput=1

\documentclass[11pt]{article}

\usepackage[final]{ACL2023}

\usepackage{times}
\usepackage{latexsym}

\usepackage[T1]{fontenc}

\usepackage[utf8]{inputenc}
\usepackage{amsmath}
\usepackage{amssymb}
\usepackage{microtype}

\usepackage{booktabs, multirow} 
\usepackage{soul}
\usepackage{xcolor,colortbl} 
\usepackage{inconsolata}
\usepackage{graphicx}
\usepackage{subcaption}

\title{Latent-Space Intervention for Cross-Lingual Factual Consistency: Consistency Improvements without Accuracy Drops}

\author{Faeze Ghorbanpour$^{1,3}$\qquad Constanza Fierro$^{2}$ \qquad Alexander Fraser$^{1, 3}$ \qquad Anders Søgaard$^{2}$ \vspace{.2cm}\\ 
$^{1}$School of Computation, Information and Technology, TU Munich \\
$^{2}$Department of Computer Science, University of Copenhagen\\ 
$^{3}$Munich Center for Machine Learning (MCML)\\
\vspace{.1cm} {\tt \small faeze.ghorbanpour@tum.de, c.fierro@di.ku.dk} 
}

\begin{document}
\maketitle
\begin{abstract}

Large Language Models (LLMs) often answer the same factual question differently across languages. We study whether cross-lingual latent-space intervention can reduce this inconsistency. We train layer-specific autoencoders on parallel multilingual representations and apply inference-time corrections to factual QA prompts. We find that latent intervention improves geometric alignment between languages, and that this improvement translates into consistent gains in cross-lingual consistency with English across both open-ended and multiple-choice QA formats, without degrading factual accuracy. In open-ended QA, Spearman's rank correlation between English and non-English languages improves substantially, with gains of 0.16 for English-Arabic and 0.20 for English-Russian pairs. In multiple-choice QA, answer agreement with English improves consistently across both KLAR and mParaRel. 
Ablations show that AE reconstruction yields consistent gains at no accuracy cost, while PCA projection contributes marginally, and mean-shift produces substantially larger consistency gains in open-ended QA at the cost of some accuracy.~\footnote{The code is available at \href{https://github.com/FaezeGhorbanpour/LatentInterventionCrosslingualFactuality}{github.com/ LatentInterventionCrosslingualFactuality}.}
\end{abstract}

\section{Introduction}



Multilingual language models (MLMs) achieve strong performance on many cross-lingual tasks~\citep{zhu2024multilingual}, but factual knowledge retrieval remains uneven across languages~\citep{jiang-etal-2020-x, Qi_2023, xing2024evaluating}. Unlike many syntactic or semantic tasks, factual recall is closely tied to the distribution of facts in pretraining data~\citep{liu2025tracing}: frequently observed facts are more likely to be recalled directly, whereas low-frequency facts often depend on cross-lingual transfer~\citep{chang2024large}. Because pretraining corpora are highly imbalanced across languages, the same fact may be well represented in one language but rare or absent in another~\citep{xu-etal-2023-language-representation}. As a result, factual knowledge can be unevenly accessible across languages~\citep{aggarwal2025language, goldman2025eclektic}. This gives rise to cross-lingual factual inconsistency: a model may answer a factual query correctly in one language while failing or producing a different answer in another~\citep{Qi_2023, lu2025paths, fierro2025multilinguallanguagemodelsremember}.

A growing body of work suggests that part of this difficulty is representational~\citep{ifergan2025beneath, lim2025language, wang-etal-2025-lost-multilinguality}. MLMs encode factual queries in representations that may entangle factual content with language-specific surface form. When these factors are not cleanly separated, the same underlying fact can occupy different regions of the model's hidden space depending on the language of the query~\citep{libovicky2019language, zhao2025less}. Recent mechanistic analyses support this view: factual knowledge appears to be represented in relatively language-independent intermediate-layer spaces, while failures can arise during the transition to later language-specific representations~\citep{wang-etal-2025-lost-multilinguality}. Other work further suggests that semantically equivalent facts across languages can activate overlapping sets of language-agnostic factual neurons, although access to these shared mechanisms is not uniform across languages~\citep{cao2026one, zhang-etal-2025-multilingual}.

This motivates a natural question: Can we construct a representation space in which factual content is more comparable across languages, and use it to test whether stronger alignment improves factual consistency? Such a space would serve two purposes. First, it would provide a diagnostic view of factual representations with reduced language-specific variation. Second, it would offer an intervention point for applying lightweight inference-time corrections without retraining the model or requiring language-specific factual supervision.

We propose that a cross-lingual autoencoder latent space, trained on parallel multilingual text, provides exactly this. By mapping LLM representations into a compact shared latent space where the same content across languages is brought into alignment, we obtain a language-agnostic view of the model's internal representations \citep{peng-etal-2025-debiasing}. Within this space, we identify a factual inconsistency subspace representing directions of maximum cross-lingual divergence for the same facts, and apply lightweight inference-time corrections via forward hooks without modifying model weights. In other words, we project out the cross-lingual, latent inconsistency subspace via PCA. 
We evaluate on KLAR \citep{wang-etal-2025-lost-multilinguality} and mParaRel \cite{fierro-sogaard-2022-factual} across typologically diverse languages in both multiple-choice and open-ended factual QA formats. Our findings show that latent-space intervention improves cross-lingual representation alignment and answer consistency without incurring a performance cost. 

\textbf{Contributions:}
\textbf{(1)} We show that autoencoder-based latent-space intervention consistently improves cross-lingual representation alignment, with larger gains for typologically distant languages and at middle-to-late layers.
\textbf{(2)} We show that this alignment improvement translates into higher semantic consistency when projecting out the dimensions along which semantic inconsistencies occur. Question-answering accuracy is largely unaffected by this intervention.

\section{Related Work}

Several works correct cross-lingual gaps without retraining. \citet{wang-etal-2025-bridging} learn alignment matrices between language representations, and \citet{kirtane2026language} apply steering vectors via forward hooks for multilingual in-context learning. A long line of work shows that shared latent spaces improve cross-lingual transfer \cite{conneau-etal-2020-unsupervised, chi-etal-2021-improving}. More directly, \citet{xu-etal-2023-language-representation} propose parameter-free projection modules that convert non-English representations into English-like equivalents to improve factual knowledge retrieval, but assume English as a fixed pivot. Our method requires no weight modification and no pivot language assumption.

Factual knowledge in MLMs is unevenly distributed across languages due to pretraining data imbalance \cite{jiang-etal-2020-x, kassner-etal-2021-multilingual}, and models remain inconsistent across languages even for paraphrases of the same fact \cite{fierro-sogaard-2022-factual}. \citet{wang-etal-2025-lost-multilinguality} provide a mechanistic account: MLMs encode facts in a language-independent concept space through most layers but fail during the final transition to language-specific representations, causing inconsistent predictions. \citet{lim2025language} and \citet{tezuka-inoue-2025-transfer} further confirm this through activations and latent process analysis. These works are diagnostic; we build on their benchmarks and metrics to evaluate a mitigation method.

\section{Our Approach}


Given a multilingual language model $\mathcal{M}$ and a set of factual relations $\mathcal{R}$, we consider a set of facts $\mathcal{F} = \{(s, r, o)\}$ where $s$ is a subject, $r \in \mathcal{R}$ is a relation, and $o$ is the correct object. Each fact is expressed as a prompt across a set of languages $\mathcal{L} = \{l_1, \ldots, l_n\}$, yielding language-specific prompts $\{x^l\}_{l \in \mathcal{L}}$ that ask the same factual question in different languages. 
We define $\mathcal{M}$ as \textit{cross-lingually inconsistent} on a fact $(s, r, o)$ if it predicts different answer choices for the same fact across languages, i.e., $\hat{y}^{l_i} \neq \hat{y}^{l_j}$ for some $l_i, l_j \in \mathcal{L}$, where $\hat{y}^l$ denotes the model's predicted choice for the prompt in language $l$. Our goal is to construct a cross-lingual latent space in which inconsistency can be measured and mitigated at inference time, without modifying model weights.

\paragraph{Cross-lingual Latent Space.}
A key challenge is that LLM representations conflate factual content with language-specific surface signals. Directly comparing or manipulating representations across languages in the LLM's native space is therefore noisy. Following \citet{peng-etal-2025-debiasing}, we address this by mapping LLM representations into a compact cross-lingual latent space where the same content across languages is brought into closer alignment.

Specifically, we train an autoencoder (AE) in two phases. First, we optimize a shared encoder and decoder jointly. Given a parallel pair $(h^{l_1}, h^{l_2})$ of LLM hidden states for semantically equivalent sentences in languages $l_1$ and $l_2$, with latent representations $z^{l_i} = \text{enc}(h^{l_i})$ and reconstructions $\hat{h}^{l_i} = \text{dec}(z^{l_i})$, the training objective is:
\noindent
\begin{equation}
\mathcal{L} = \zeta\,\mathcal{L}_{\text{self}} + 
\beta\,\mathcal{L}_{\text{align}} + 
\eta\,\mathcal{L}_{\text{var}} + 
\rho\,\mathcal{L}_{\text{cov}}
\label{eq:loss}
\end{equation}

\noindent where the self-reconstruction loss
\begin{equation}
\mathcal{L}_{\text{self}} = \ell(\hat{h}^{l_1}, h^{l_1}) + \ell(\hat{h}^{l_2}, h^{l_2})
\end{equation}
encourages faithful reconstruction within each language.
The alignment loss
\noindent
\begin{equation}
\mathcal{L}_{\text{align}} = \text{MSE}(\tilde{z}^{l_1}, \tilde{z}^{l_2}), \quad \tilde{z}^{l_i} = \frac{z^{l_i}}{\|z^{l_i}\|}
\end{equation}
directly minimizes the distance between normalized latent representations of parallel pairs. The variance loss
\noindent
\begin{equation}
\mathcal{L}_{\text{var}} = \sum_j \text{ReLU}\!\left(\gamma - \text{Std}_j(Z)\right)
\end{equation}
prevents dimensional collapse by penalizing dimensions whose standard deviation falls below a threshold $\gamma$. The covariance loss
\noindent
\begin{equation}
\mathcal{L}_{\text{cov}} = \frac{1}{k}\sum_{i \neq j} \left[\text{Cov}(Z)\right]^2_{ij}
\end{equation}
reduces redundancy by penalizing off-diagonal entries of the latent covariance matrix. We use the Huber \citep{huber1992robust} loss for $\ell$. Ablating the self-reconstruction and alignment terms confirms that each is necessary: removing self-reconstruction collapses the latent space, while removing alignment substantially weakens cross-lingual retrieval (Appendix~\ref{sec:loss_ablation}).

In the second phase, we freeze the shared encoder and train language-specific decoders $\{\text{dec}_l\}_{l \in \mathcal{L}}$ independently, one per language, minimizing $\ell(\text{dec}_l(z^l), h^l)$ for each language $l$. This allows each decoder to specialize in reconstructing its target language's representations from the shared language-agnostic latent space, while preserving the cross-lingual alignment learned in phase one. 
Training the encoder and language-specific decoders jointly instead degrades both consistency and accuracy at every intervention layer, confirming that the two-phase split is necessary (Appendix~\ref{sec:training_ablation}).


\paragraph{Inference-time Correction.}
At inference time, we register a forward hook on a chosen transformer layer. When the model processes a prompt, the hook intercepts the hidden states $\mathbf{X} \in \mathbb{R}^{B \times T \times d}$, computes a mean-pooled sentence representation $\mathbf{h} = \frac{1}{T}\sum_{t=1}^{T} \mathbf{x}_t \in \mathbb{R}^d$, and maps it to the latent space via the AE encoder: $\mathbf{z} = \text{enc}(\mathbf{h})$. We explore three correction strategies in the latent space.

The \textit{AE} variant applies no correction; it simply encodes and decodes the representation without modification, serving as a baseline that isolates the effect of the AE bottleneck itself from any explicit correction:   $ \tilde{\mathbf{z}} = \mathbf{z}$.
The \textit{projection} variant identifies an inconsistency subspace via PCA on cross-lingual difference vectors 
and projects it out:
\noindent
\begin{equation}
    \tilde{\mathbf{z}}^l = \mathbf{z}^l - \sum_{j=1}^{m} (\mathbf{z}^l \cdot \mathbf{d}_j)\, \mathbf{d}_j
\end{equation}
where $\mathbf{d}_1, \ldots, \mathbf{d}_m$ are the top-$m$ principal components of the inconsistency subspace.

The \textit{mean-shift} variant pulls each language's representation toward the cross-lingual mean without assuming any pivot language:
\noindent
\begin{equation}
    \tilde{\mathbf{z}}^l = \mathbf{z}^l + \lambda\left(\bar{\mathbf{z}} - \bar{\mathbf{z}}^l\right)
\end{equation}
where $\bar{\mathbf{z}}^l = \frac{1}{N}\sum_{i=1}^{N} \mathbf{z}^l_i$ is the mean representation of language $l$ across all $N$ facts in our evaluation sets and $\bar{\mathbf{z}} = \frac{1}{|\mathcal{L}|}\sum_{l \in \mathcal{L}} \bar{\mathbf{z}}^l$ is the mean across all languages.

In all three cases, the corrected latent vector $\tilde{\mathbf{z}}$ is decoded back to the LLM representation space via the language-specific decoder $\text{dec}_l(\tilde{\mathbf{z}})$, and the difference between the decoded representation and the original mean-pooled hidden state is added as a residual to all token positions:
\noindent
\begin{equation}
    \mathbf{x}'_t = \mathbf{x}_t + \bigl(\text{dec}_l(\tilde{\mathbf{z}}) - \mathbf{h}\bigr), \quad \forall t \in \{1, \ldots, T\}
\end{equation}

\section{Experimental Setup}


\paragraph{Datasets}

For autoencoder training, we use parallel sentences from TED talk scripts \citep{salesky2021multilingualtedxcorpusspeech}, and evaluate latent space alignment quality on \textit{FLORES+} \citep{maillard-etal-2024-findings, goyal-etal-2022-flores}. We use a subset of five typologically diverse languages: Arabic, English, Dutch, Russian, and Chinese. 
We evaluate on two multilingual factual knowledge benchmarks. \textit{KLAR} \cite{wang-etal-2025-lost-multilinguality} is a cross-lingual factual consistency dataset, with prompts available in 17 languages.
\textit{mParaRel} \cite{fierro-sogaard-2022-factual} is a multilingual extension of ParaRel \cite{Elazar2021MeasuringAI} containing paraphrase sets for relational facts across 45 languages. For both datasets, we convert facts into multiple-choice questions with four candidates, one correct answer, and three distractors drawn from the same relation, ensuring a consistent evaluation format across datasets and languages. We do not evaluate open-ended QA on mParaRel, as its cloze-style prompts are designed for masked language models and do not place the answer at the end. For details on the datasets, see Appendix \ref{sec:data}.

\paragraph{Models.}
We evaluate three open-source 8B-parameter models spanning different multilingual pretraining regimes: Aya-expanse-8B~\cite{dang2024ayaexpansecombiningresearch}, an instruction-tuned multilingual model covering 23 languages; Llama-3.1-8B~\citep{grattafiori2024llama3herdmodels}, a predominantly English-centric model; and Qwen-3-8B~\citep{yang2025qwen3technicalreport}. This spread lets us test whether the effect of latent-space intervention depends on how well aligned a model already is across languages. For each model, we train separate layer-specific autoencoders following the procedure in Section 3. Appendix~\ref{sec:details} provides details about the models and hyperparameters.



\paragraph{Evaluation Metrics.}
\label{sec:setup}
We evaluate both the learned latent space and the downstream factual interventions. First, on FLORES+ parallel sentences, we assess the autoencoder before and after reconstruction using t-SNE, cosine similarity, and parallel-sentence retrieval accuracy.
For factual QA, we report three metrics. \textit{Cosine similarity to English} measures representation-level alignment between each non-English language and English. \textit{Factual accuracy} measures the proportion of examples answered correctly: for open-ended QA, whether the generated answer matches the gold object; for multiple-choice QA, whether the selected candidate is correct. \textit{Agreement} for multiple-choice QA measures how often two languages produce the same answer, regardless of whether that answer is correct. 
For open-ended QA, the model generates an answer autoregressively, and we measure cross-lingual consistency as the Spearman rank correlation between the rank assigned to the correct answer and the ranks assigned to the distractors (Rank Corr).

\paragraph{Choice of intervention layer.} Our layer-wise alignment analysis (Appendices~\ref{sec:alignment} and~\ref{sec:results}) shows that raw representations of parallel content are already partially mixed at middle layers but separate into language-specific clusters toward the end of the network, and that AE-based correction has the largest effect precisely where this separation is strongest. We therefore intervene at middle-to-late layers, and report results at the best-performing layer per setting. This choice transfers across models: the most effective layers fall in the final third of the network for all three model families we evaluate (L20–L28 for Aya-expanse-8B, L20–L24 for Llama-3.1-8B, and L28–L32 for Qwen-3-8B), consistent with the mechanistic account of \citet{wang-etal-2025-lost-multilinguality}, in which factual recall diverges into language-specific form during the final transition.

\begin{table*}[t]
\centering
\small
\begin{tabular}{llcccccc}
\toprule
& & \multicolumn{2}{c}{KLAR OE} & \multicolumn{2}{c}{KLAR MC} & \multicolumn{2}{c}{mParaRel MC} \\
\cmidrule(lr){3-4} \cmidrule(lr){5-6} \cmidrule(lr){7-8}
& Intervention & Acc & Rank Corr (en) & Acc & Agr (en) & Acc & Agr (en) \\
\midrule
\multirow{3}{*}{Raw}
& No intervention        & 56.77 & 6.68 & 85.59 & 85.09 & 85.91 & 86.70 \\
& + Mean-shift & 53.78 & 14.09 & 85.59 & 85.01 & 81.09 & 76.02 \\
& + PCA        & 56.72 & 6.81 & 85.56 & 84.97 & 85.86 & 86.29 \\
\midrule
\multirow{3}{*}{AE}
&  No intervention       & 56.40 & 7.77 & 85.32 & \textbf{85.43} & \textbf{86.35} & 86.97 \\
& + Mean-shift & 53.45 & \textbf{24.91} & 85.42 & 85.41 & 85.91 & 86.68 \\
& + PCA        & 56.30 & 8.25 & 85.29 & \textbf{85.43} & 86.37 & \textbf{86.97} \\
\bottomrule
\end{tabular}
\caption{Average accuracy (\%) and consistency with English across strategies at the best layer per setting. Rank Corr (en) is the average Spearman rank correlation ($\times 100$) across English-paired language pairs. Agr (en) is the average answer agreement (\%) with English. Bold indicates best result per column.}
\label{tab:main_results}
\end{table*}

\begin{table*}[t]
\centering
\small
\begin{tabular}{lcccccccccccc}
\toprule
& \multicolumn{4}{c}{KLAR OE Rank Corr} & \multicolumn{4}{c}{KLAR MC Agreement} & \multicolumn{4}{c}{mParaRel MC Agreement} \\
\cmidrule(lr){2-5} \cmidrule(lr){6-9} \cmidrule(lr){10-13}
& ar-en & en-nl & en-ru & en-zh & ar-en & en-nl & en-ru & en-zh & ar-en & en-nl & en-ru & en-zh \\
\midrule
Raw            & 2.52 & 16.21 & 4.55 & 3.45 & 66.17 & 94.65 & 90.87 & 88.66 & 84.31 & 90.06 & 87.33 & 85.09 \\
AE       & 2.51 & 17.83 & 4.93 & 5.82 & 66.36 & \textbf{95.23} & \textbf{91.22} & \textbf{88.89} & \textbf{84.80} & 90.55 & \textbf{87.43} & 85.19 \\
+ PCA          & 3.77 & 18.52 & 5.85 & 4.84 & 66.44 & 95.19 & 91.22 & 88.85 & \textbf{84.80} & 90.45 & \textbf{87.43} & 85.19 \\
+ Mean-shift   & \textbf{18.17} & \textbf{38.27} & \textbf{24.35} & \textbf{18.86} & \textbf{66.82} & \textbf{95.23} & 91.18 & 88.81 & 84.30 & \textbf{90.90} & 86.71 & \textbf{85.80} \\
\bottomrule
\end{tabular}
\caption{Pairwise English consistency before and after intervention at the best layer per setting. Rank Corr values are Spearman rank correlation ($\times 100$). Agreement values are answer agreement (\%).}
\label{tab:pairwise_en}
\end{table*}

\section{Results}
Across all layers and languages, AE-based intervention consistently improves cross-lingual representation alignment, with the largest gains for typologically distant languages and at later layers where raw representations become increasingly language-specific. A full evaluation of latent space alignment quality is provided in Appendices~\ref{sec:flores} and \ref{sec:alignment}.

Tables~\ref{tab:main_results} and~\ref{tab:pairwise_en} report our main results with the Aya-expanse-8B model. We organize findings around three questions: whether the intervention improves consistency with English, whether factual accuracy is preserved, and which intervention strategy drives the gains.
In open-ended QA, Spearman's rank correlation with English improves under every AE-based strategy. AE-only intervention raises average rank correlation with English from 6.68 to 7.77, and AE+PCA to 8.25. The largest gain comes from AE+Mean-shift, which raises average rank correlation to 24.91 and improves every English-paired language: English-Arabic (2.52 $\rightarrow$ 18.17), English-Dutch (16.21 $\rightarrow$ 38.27), English-Russian (4.55 $\rightarrow$ 24.35), and English-Chinese (3.45 $\rightarrow$ 18.86). In multiple-choice QA, answer agreement with English improves consistently across both datasets. For KLAR, the average English-pair agreement rises from 85.09 to 85.43. For mParaRel, it rises from 86.70 to 86.97. While gains in multiple-choice are modest, they are consistent across all English-paired language pairs.

Factual accuracy remains stable under the AE-based strategies. In multiple-choice QA, accuracy changes by less than 0.3 points in all conditions. In open-ended QA, AE and AE+PCA leave accuracy essentially unchanged (56.77 $\rightarrow$ 56.40 and 56.30), confirming that their consistency gains do not come at the cost of factual performance. AE+Mean-shift is the exception: its larger consistency gain is accompanied by a drop in average accuracy from 56.77 to 53.45, with the largest declines for Dutch and Russian.
The two effects separate cleanly by strategy. Encoding and decoding through the AE, with or without PCA projection, yields small but cost-free consistency improvements, indicating that the cross-lingual alignment induced by the AE mapping is sufficient to improve consistency without disturbing factual content. Mean-shift, which additionally translates each language toward the cross-lingual mean, produces substantially larger consistency gains but perturbs representations enough to degrade answer accuracy. This exposes a tradeoff between cross-lingual consistency and factual accuracy that the weaker interventions do not reveal, and reinforces our broader observation that representation alignment and factual consistency are related but distinct.

To test whether these findings depend on the choice of model, we repeat the KLAR evaluation on Llama-3.1-8B and Qwen-3-8B (Appendix~\ref{sec:other_models}). AE-based intervention improves average consistency with English on both models and in both formats, with the largest gains on Llama-3.1-8B, whose baseline cross-lingual agreement is lowest, and smaller gains on Qwen-3-8B, where baseline agreement is already high. The accuracy cost of mean-shift is confined to open-ended generation and does not appear in multiple-choice selection, consistent with factual recall in open-ended settings being more sensitive to representational intervention than structured answer selection.


\section{Conclusion}

We investigated whether inference-time intervention in a cross-lingual latent space can reduce factual inconsistency in MLMs. Our results across KLAR \cite{wang-etal-2025-lost-multilinguality} and mParaRel \cite{fierro-sogaard-2022-factual} show that AE-based intervention improves cross-lingual representation alignment, particularly for typologically distant languages and at middle-to-late layers, and translates into improved cross-lingual consistency. 

\section*{Limitations}


Our study has several limitations that also point to useful directions for future work. First, we evaluate on a focused set of languages and factual relations from KLAR, as well as on mParaRel. This controlled setting allows us to analyze cross-lingual factual inconsistency in detail, but broader evaluation on additional benchmarks, languages, and relation types is needed to test how general the observed patterns are.

Second, our experiments span three model families at a single parameter scale (8B). Evaluating across Aya-expanse-8B, Llama-3.1-8B, and Qwen-3-8B establishes that the main patterns are not tied to a particular pretraining mixture, and that the effective intervention layers fall consistently in the final third of the network across all three. Extending this analysis to larger scales would test how far this depth heuristic generalizes, and would clarify how the available headroom for correction relates to a model's baseline cross-lingual alignment.

Third, our interventions are intentionally lightweight and inference-time only. PCA-based removal and mean-shift correction provide simple and interpretable ways to manipulate latent representations, but factual inconsistency may be distributed across relation-specific or nonlinear directions. More expressive correction methods may therefore yield stronger factual gains while preserving the interpretability of the current framework.

Finally, our results show that latent-space intervention improves geometric alignment more reliably than factual accuracy or agreement. Rather than treating this as a failure, we view it as an important finding: representation alignment is not sufficient by itself to guarantee factual consistency. Future work can build on this distinction by developing interventions that explicitly preserve answer-relevant information while reducing cross-lingual divergence.


\section*{Ethical Considerations}

This work uses publicly available multilingual benchmarks and openly licensed models, and involves no human subjects or personal data. Our method operates entirely at inference time and leaves model weights unchanged, so it is straightforward to audit, disable, or apply selectively, and it introduces no new training data or supervision. We evaluate factual accuracy alongside consistency throughout, precisely because consistency alone is not a sufficient objective: an intervention that aligns answers across languages should be shown to preserve correctness as well. Reporting both together is what allows the method to be assessed responsibly, and we release our code and trained models to support such assessment.

\section*{Acknowledgements}
The work was supported by the European Research Council (ERC) through the European Union's Horizon Europe research and innovation program (grant agreement No. 101113091) and the German Research Foundation (DFG; grant FR 2829/7-1). CF was supported by Danish Data Science Academy, which is funded by the Novo Nordisk Foundation (NNF21SA0069429). 

The authors gratefully acknowledge the scientific support and HPC resources provided by the Erlangen National High Performance Computing Center (NHR@FAU) of the Friedrich-Alexander-Universität Erlangen-Nürnberg (FAU) under the NHR project b279bb. NHR funding is provided by federal and Bavarian state authorities. NHR@FAU hardware is partially funded by the German Research Foundation (DFG) – 440719683.

\bibliography{custom}
\bibliographystyle{acl_natbib}

\appendix

\nocite{ghorbanpour-etal-2025-data}

\section{Data Details}
\label{sec:data}
We train the AE on parallel multilingual sentence pairs sourced from TED Talk transcripts~\citep{salesky2021multilingualtedxcorpusspeech} across five languages: Arabic (ar), English (en), Dutch (nl), Russian (ru), and Chinese (zh). After deduplication, each language contains 212,698 unique sentences, yielding 2,126,980 parallel pairs across all ten language combinations. The data is split into train, development, and test sets. For evaluation of the cross-lingual latent space, we use 500 parallel sentences per language from the FLORES+ benchmark~\citep{maillard-etal-2024-findings, goyal-etal-2022-flores}, using 500 sentences per language from the devtest split.

For KLAR~\cite{wang-etal-2025-lost-multilinguality}, we use 2,619 facts spanning 20 factual relations (e.g.\ capital cities, places of birth, nationalities) across five languages: Arabic, English, Dutch, Russian, and Chinese. For each fact, one prompt template is selected and kept consistent across all languages to ensure comparable evaluation conditions.

For mParaRel~\cite{fierro-sogaard-2022-factual}, we use 1,026 facts spanning 24 relations across the same five languages. As with KLAR, one template per fact is randomly selected and held fixed across languages. For both datasets, each fact is paired with four answer candidates, one
correct object, and three distractors drawn from the same relation. The
candidate set supports both the multiple-choice format and the rank
correlation metric described in Section~\ref{sec:setup}.

\section{Experimental Details}
\label{sec:details}
The AE encoder and decoders are three-layer MLPs with a latent dimension of $k=256$. We train using the AdamW optimizer with a learning rate of $10^{-4}$ and weight decay of $10^{-8}$, and a ReduceLROnPlateau scheduler with a reduction factor of $0.5$, patience of $5$, and minimum learning rate of $10^{-7}$. Both training phases use a batch size of $1024$, a maximum of $128$ epochs, early stopping with patience of $16$, and a maximum sequence length of $128$ tokens. The loss weights are set to $\zeta=1$, $\beta=0.1$, $\eta=1$, $\rho=1$, and $\gamma=1$. We train separate autoencoders for each transformer layer, using mean-pooled hidden states extracted from that layer. At inference time, the forward hook is registered on each transformer layer output, using mean pooling over token positions. The correction is applied only during the prefill step and skipped for autoregressive generation steps. 
We evaluate three correction strategies, \textit{AE} (encode-decode only), \textit{PCA projection}, and \textit{mean-shift}, with $\lambda = 0.6$ and $m = 1$ principal component for PCA. For open-ended QA, answers are generated greedily with a maximum of five new tokens. Because the models differ in depth,
we sweep a per-model layer grid: $\{4, 8, 12, 16, 20, 24, 28, 31\}$ for
Aya-expanse-8B and Llama-3.1-8B (32 layers), and
$\{4, 8, 12, 16, 20, 24, 28, 32, 35\}$ for Qwen-3-8B (36 layers).

All experiments were conducted on a single NVIDIA A100 GPU (40/80 GB). We used the HuggingFace Transformers library \citep{wolf-etal-2020-transformers} with bf16 precision throughout. For each of the 8 evaluated layers, we extracted hidden-state representations from each model, trained a separate autoencoder, and trained language-specific decoders independently, totaling approximately 5 hours per layer. The autoencoder encoder and decoder are three-layer MLPs with hidden dimensions 1024, 512, and 256. In total, training across all layers required approximately 32 hours of GPU time.

\section{Use of Artifacts and AI Statement}

All artifacts used in this work
are openly available and licensed for research use. We use artifacts in accordance with their intended purpose and license terms. This work does not involve human subjects, crowd sourcing, or the collection of personal data.
We used AI assistance in several stages of this work. Claude (Anthropic) was used as a writing assistant to help draft, revise, and improve the clarity of the paper text. All AI-generated text was reviewed, edited, and verified by the authors. All experimental results, analyses, and scientific claims are the sole responsibility of the authors.

\section{Latent Space Alignment on FLORES+}
\label{sec:flores}
\begin{figure}[t]
    \centering

    \includegraphics[width=0.24\textwidth]{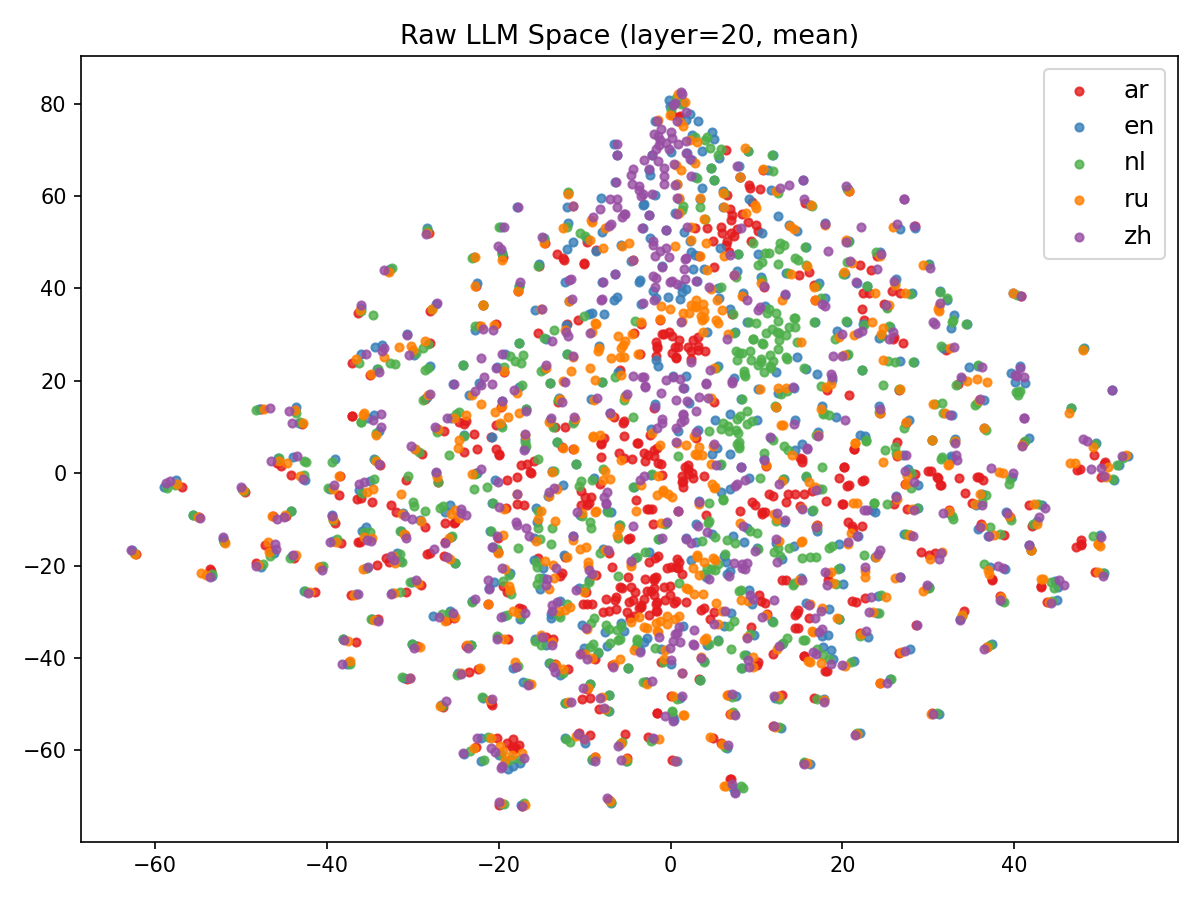}%
    \includegraphics[width=0.24\textwidth]{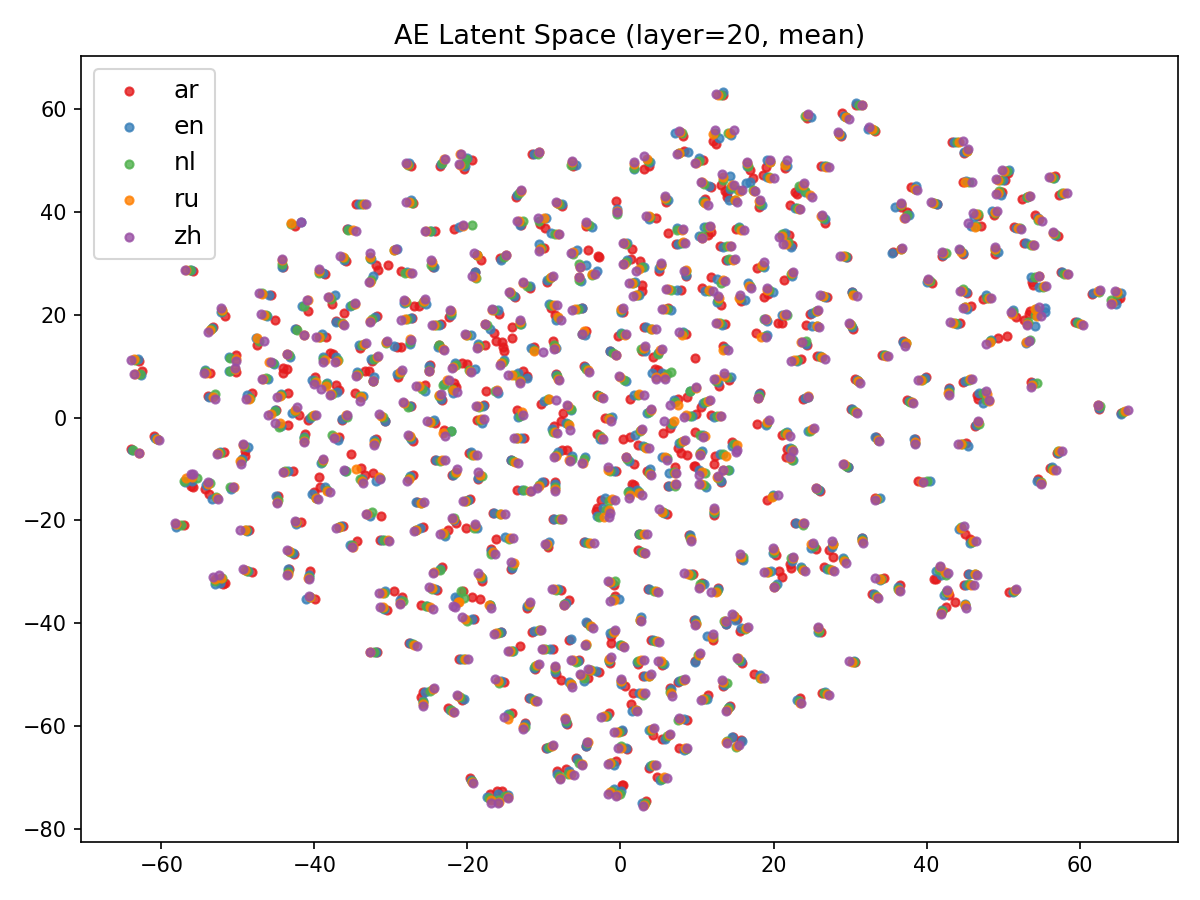}\\[-0.2em]
    \includegraphics[width=0.24\textwidth]{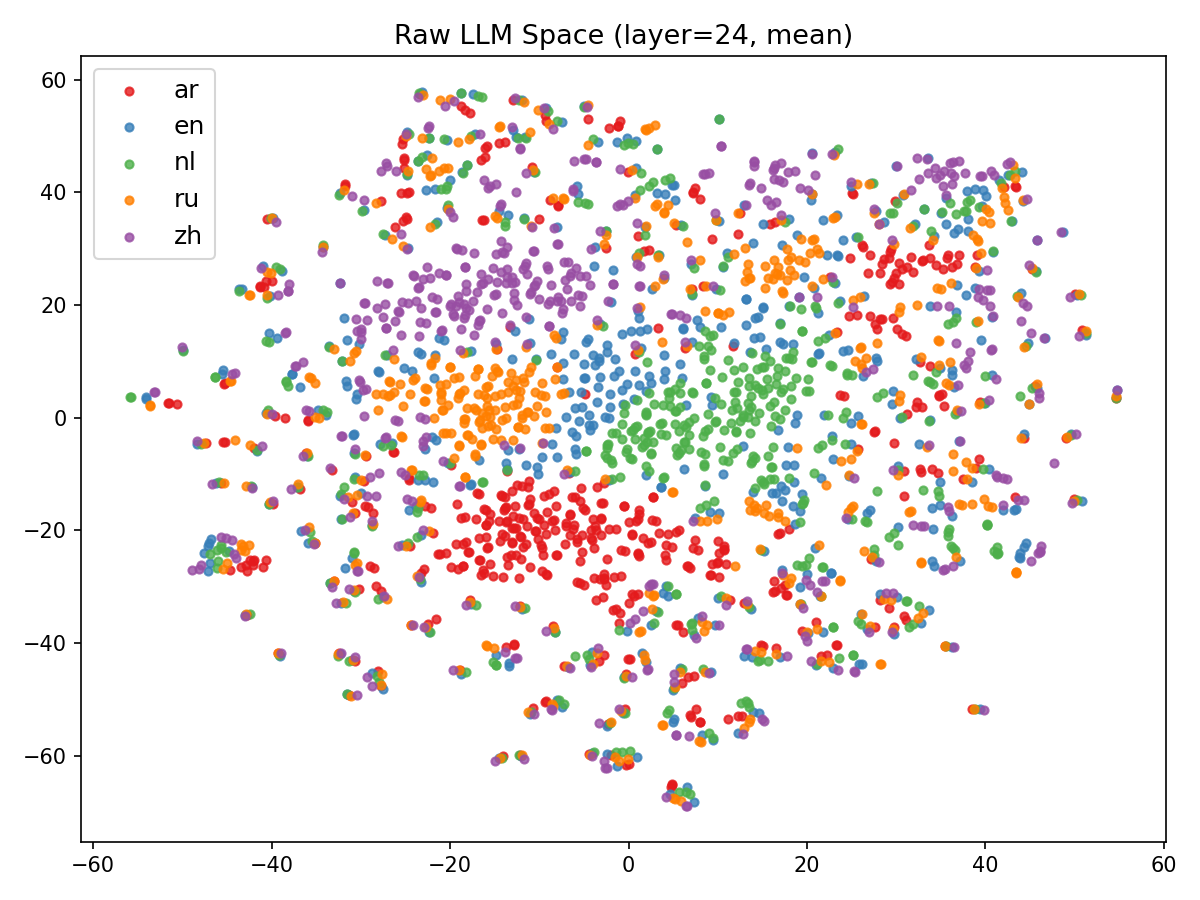}%
    \includegraphics[width=0.24\textwidth]{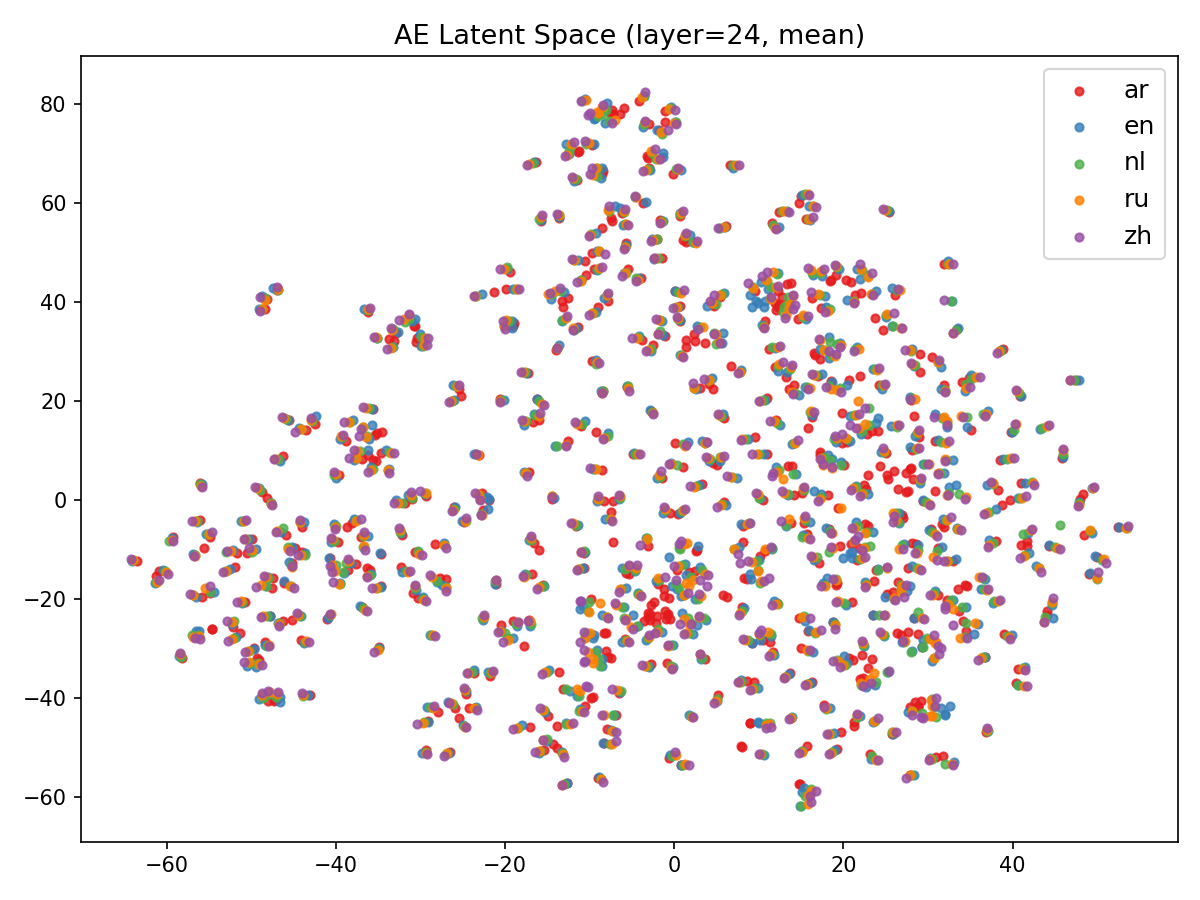}\\[-0.2em]
    \includegraphics[width=0.24\textwidth]{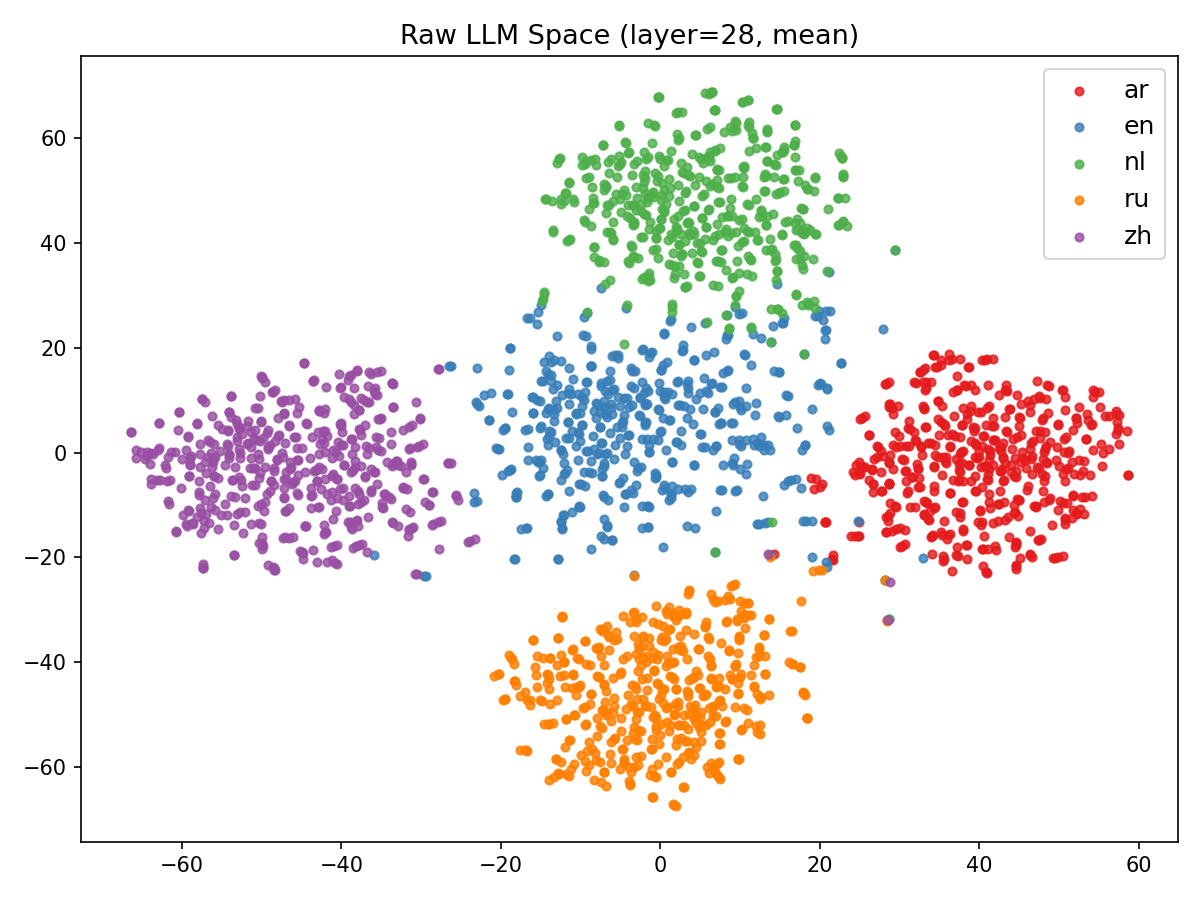}%
    \includegraphics[width=0.24\textwidth]{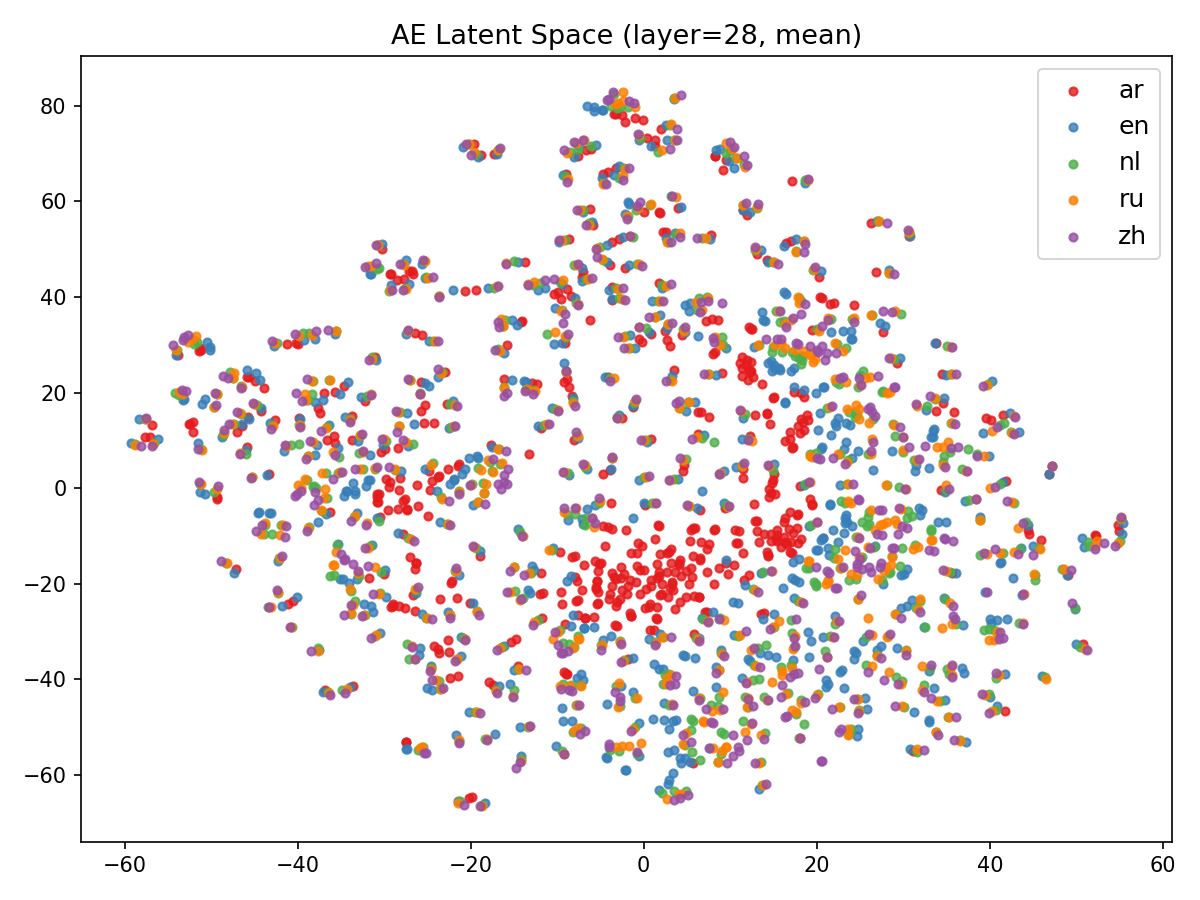}

    \caption{t-SNE visualizations of FLORES+ parallel sentence representations in the raw LLM space (left) and AE latent space (right) at layers 20, 24, and 28. Colors indicate languages.}
    \label{fig:tsne}
\end{figure}

Figure~\ref{fig:tsne} shows t-SNE visualizations of FLORES+ parallel sentence representations \citep{goyal-etal-2022-flores} in the raw LLM space and the AE latent space at layers 20, 24, and 28. In the raw LLM space, language-specific clustering becomes increasingly pronounced at deeper layers, with languages forming nearly fully separated clusters at L28. After mapping to the AE latent space, representations of parallel sentences across all five languages collapse into a single shared point cloud, confirming that the autoencoder successfully removes language-specific surface signals.

At L20, raw representations are already partially mixed across languages, and the AE provides modest gains in P1 (+0.038) with negligible cosine change, suggesting that at this layer the cross-lingual structure is partially present in the raw space and the AE refines rather than reconstructs it. At L24, language separation becomes more visible in the raw space, and the AE produces a clearer improvement in cosine similarity (+0.069), indicating that deeper layers require stronger correction to achieve alignment. At L28, where raw representations form fully separated language clusters, the AE achieves the largest gains in both cosine similarity (+0.150) and P1 (+0.042), demonstrating that the latent space can recover meaningful cross-lingual alignment even when the raw LLM space has become strongly language-specific. Together, these results confirm that the AE latent space provides consistent cross-lingual alignment across layers, with greater impact where language-specific signals are strongest.

\begin{figure*}[t]
    \centering

    \includegraphics[width=0.32\textwidth]{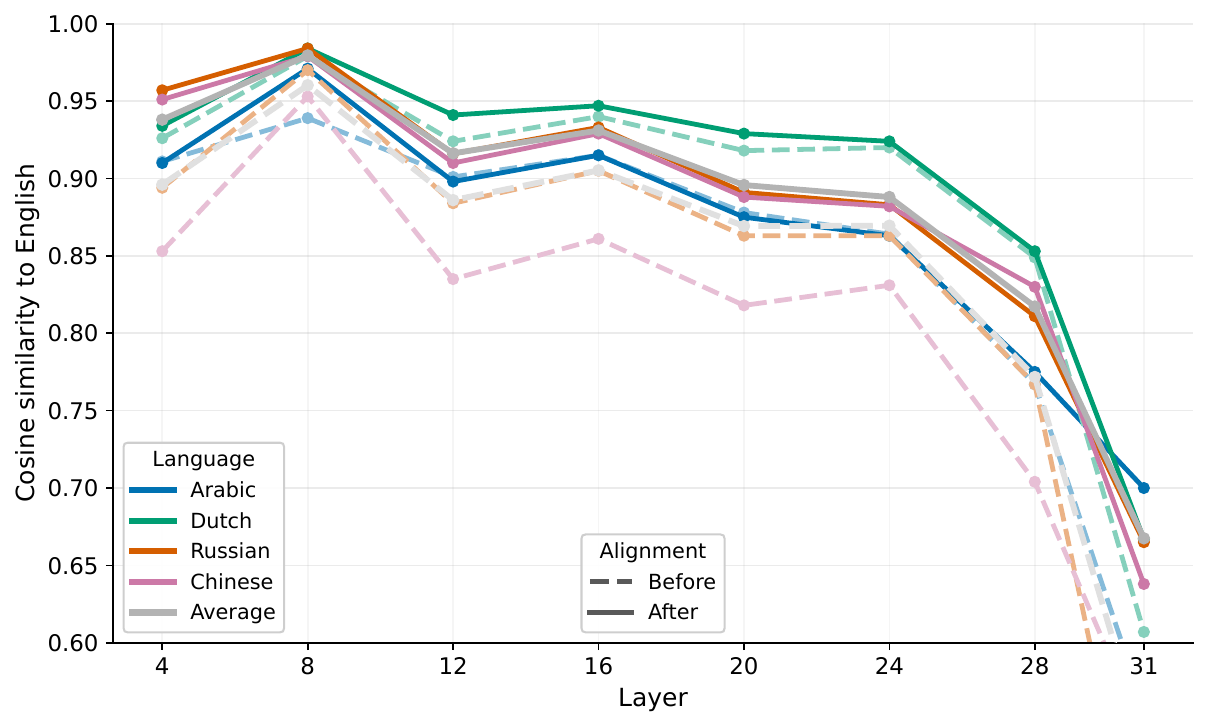}%
    \includegraphics[width=0.32\textwidth]{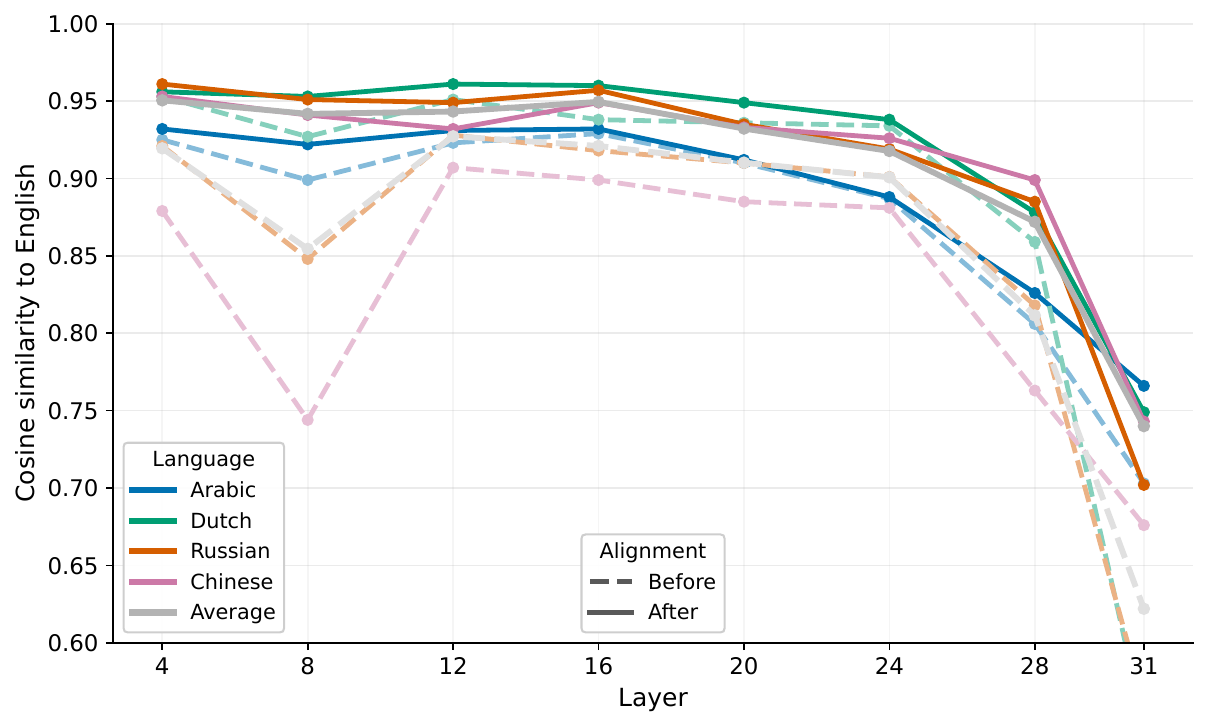}
    \includegraphics[width=0.32\textwidth]{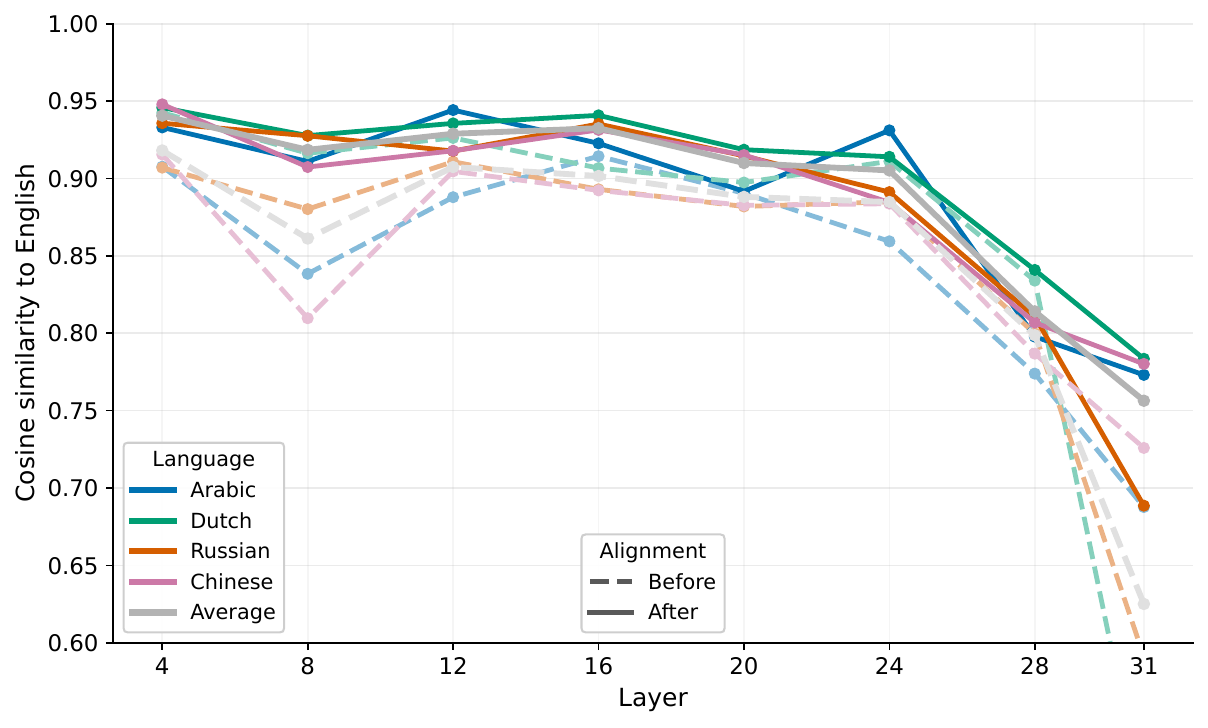}\\[-0.2em]
    \includegraphics[width=0.32\textwidth]{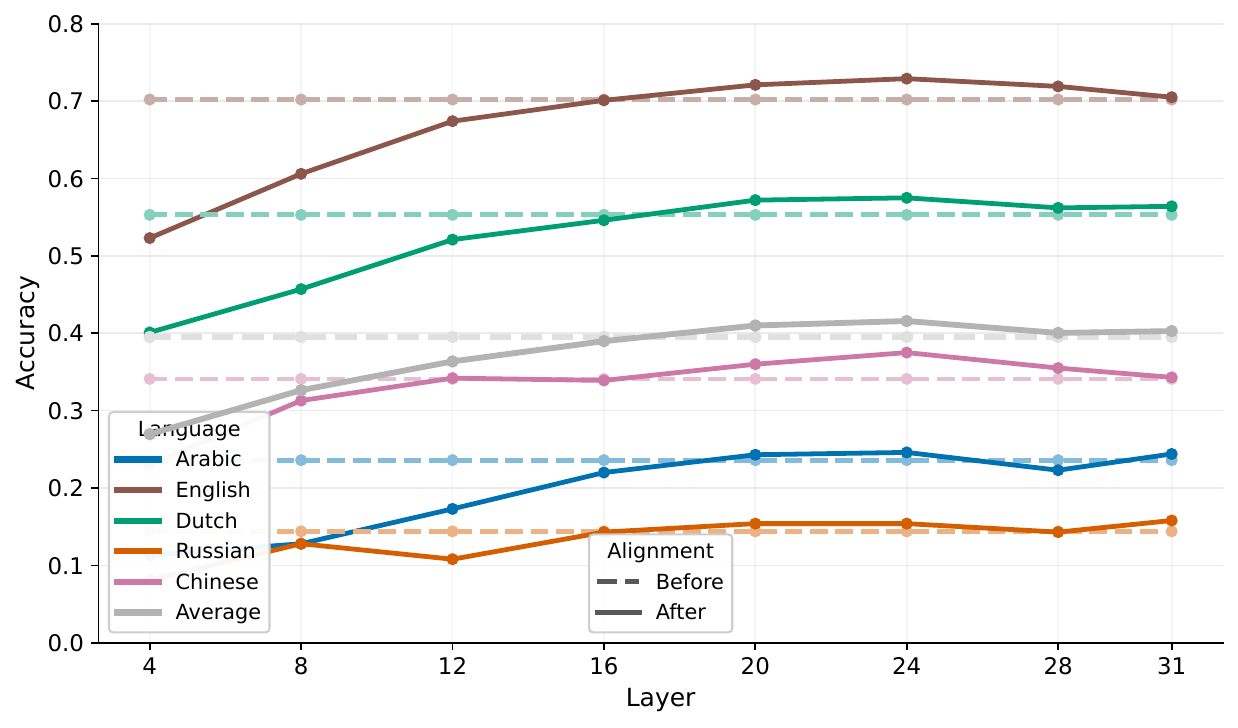}%
    \includegraphics[width=0.32\textwidth]{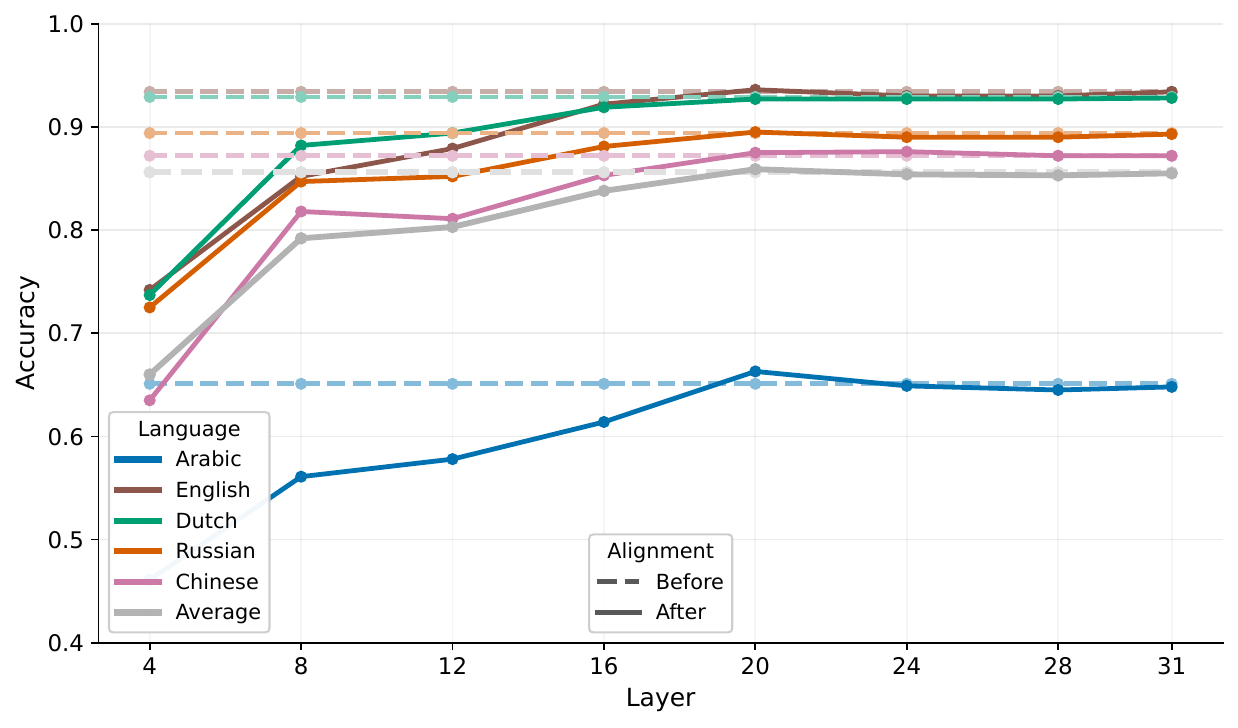}
    \includegraphics[width=0.32\textwidth]{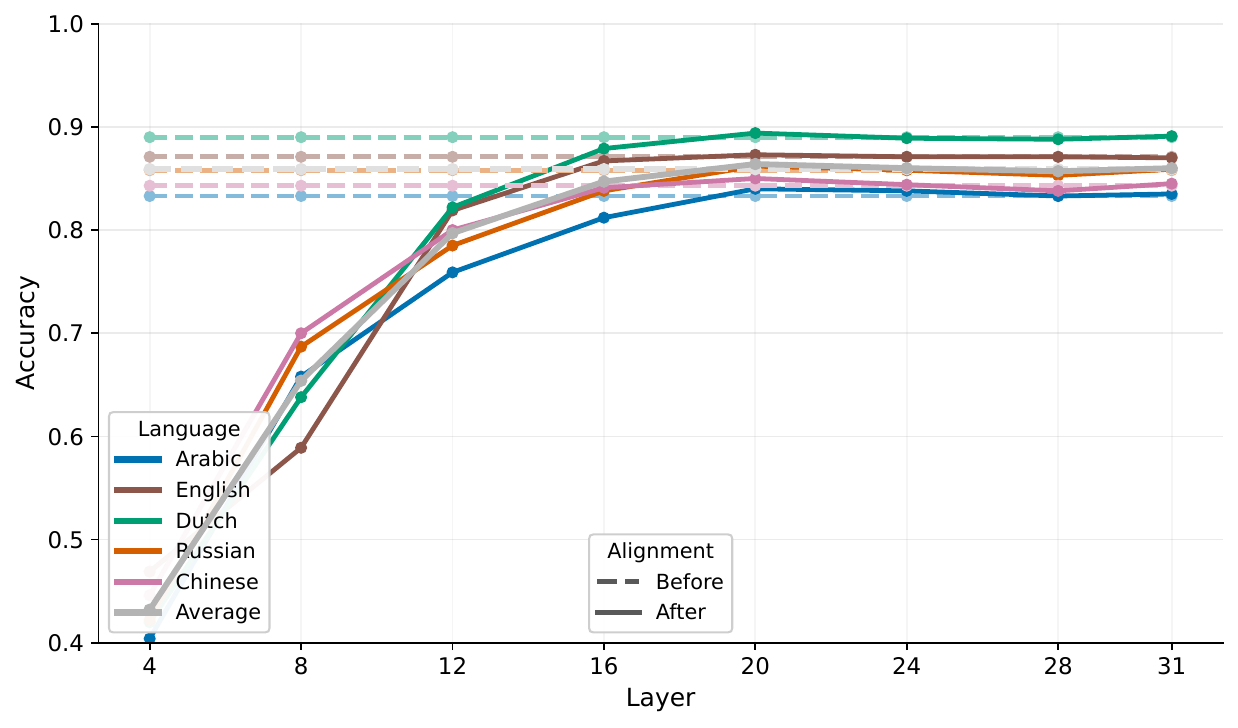}

    \caption{First row: Cosine similarity to English before and after latent-space intervention across layers for open-ended and multiple-choice QA (KLAR and mParaRel). Second row: First-token accuracy (\%) before and after latent-space intervention across layers for open-ended and Multiple-choice QA (KLAR and mParaRel). 
    Dashed lines show before intervention; solid lines show after intervention.
    }
    \label{fig:four_images}
\end{figure*}

\section{Cross-lingual Latent Space Alignment}
\label{sec:alignment}

The first row of Figure \ref{fig:four_images} shows cosine similarity to English before and after intervention across layers for open-ended and multiple-choice formats, respectively. Latent-space intervention consistently improves cross-lingual alignment across all languages and layers, with the gap between before and after being most pronounced at later layers, where unintervened representations diverge sharply from English. The improvement is largest for Chinese and Arabic, the most typologically distant languages in our set, while Dutch shows the smallest gain, consistent with its closer typological proximity to English. At middle layers (L8--L24), post-intervention similarities converge across languages, suggesting the AE latent space successfully reduces language-specific variation in this range. At the final layers (L28--L31), alignment degrades even after intervention, indicating that the AE cannot fully compensate for the strong language-specific signals that emerge in the model's final representations.

\section{Factual Accuracy}
\label{sec:results}

Bottom row of Figure~\ref{fig:four_images} shows first-token accuracy before and after AE intervention across layers for open-ended and multiple-choice QA, respectively. For open-ended QA, intervention at early layers (L4--L8) substantially degrades accuracy across all languages, with the largest drops observed for English and Dutch. As the intervention layer increases, accuracy recovers and at middle-to-late layers (L20--L24) most languages modestly exceed their baseline. English and Dutch show the clearest gains, while Arabic remains at or below baseline throughout, and Russian shows negligible improvement. These results suggest that latent-space intervention can improve open-ended factual recall, but only when applied at the right layer. The next plot, which is for multiple-choice QA, shows a similar pattern of early-layer degradation, with accuracy recovering toward baseline at middle-to-late layers. However, unlike open-ended QA, no language consistently exceeds its baseline after intervention. Multiple-choice accuracy is already high, leaving little room for improvement, and the intervention neither meaningfully helps nor hurts at middle-to-late layers. The contrast with open-ended QA suggests that factual recall in the open-ended setting is more sensitive to representational intervention than structured answer selection.

\section{Loss Ablation}
\label{sec:loss_ablation}

The training objective in Equation~\ref{eq:loss} combines self-reconstruction, alignment, variance, and covariance terms. To verify that these components behave as intended, we retrain the autoencoder with individual terms removed and evaluate the resulting latent space on FLORES+ parallel sentences using cosine similarity between parallel pairs and parallel-sentence retrieval accuracy (P@1). Table~\ref{tab:loss_ablation} reports results on Llama-3.1-8B.

Each term contributes as designed. Removing the self-reconstruction loss causes the latent space to collapse: retrieval accuracy falls to 0.84, indicating that the space retains almost no recoverable content, and cosine similarity falls well below the raw LLM representations. Removing the alignment loss yields a functional but substantially weaker space, with retrieval accuracy dropping from 95.46 to 83.22. Cross-lingual alignment is therefore not an emergent byproduct of reconstruction: it requires explicit supervision. The full objective outperforms the raw LLM space on both measures by a wide margin.

\begin{table}[t]
\centering
\small
\begin{tabular}{lcc}
\toprule
Variant & Cosine & P@1 \\
\midrule
Raw LLM space & 67.61 & 38.36 \\
w/o self-reconstruction loss & 13.92 & 0.84 \\
w/o alignment loss & 76.02 & 83.22 \\
Full objective & \textbf{86.51} & \textbf{95.46} \\
\bottomrule
\end{tabular}
\caption{Loss ablation on FLORES+ with Llama-3.1-8B. Cosine is the average cosine similarity between parallel sentence pairs; P@1 is parallel-sentence retrieval accuracy (\%).}
\label{tab:loss_ablation}
\end{table}

\begin{table*}[t]
\centering
\small
\begin{tabular}{lcccccc|cccccc}
\toprule
& \multicolumn{6}{c|}{Open-ended QA} & \multicolumn{6}{c}{Multiple-choice QA} \\
\cmidrule(lr){2-7} \cmidrule(lr){8-13}
Method & ar-en & en-nl & en-ru & en-zh & Avg & Acc & ar-en & en-nl & en-ru & en-zh & Avg & Acc \\
\midrule
Raw & 21.70 & 38.55 & 19.46 & 6.84 & 21.64 & 45.85 & 48.76 & 89.58 & 87.06 & 79.84 & 76.31 & 78.21 \\
PCA & 24.27 & 37.95 & 19.31 & 7.31 & 22.21 & 45.48 & 48.76 & 89.69 & 87.40 & 79.92 & 76.44 & 78.25 \\
Mean-shift & 22.98 & 36.93 & 20.75 & 4.32 & 21.25 & 42.31 & 50.63 & 88.85 & 85.76 & 80.26 & 76.38 & 78.21 \\
AE & 20.71 & 40.29 & 20.96 & \textbf{8.84} & 22.70 & \textbf{49.00} & 50.13 & \textbf{89.77} & 87.74 & 80.41 & 77.01 & 78.55 \\
+ PCA & \textbf{24.48} & \textbf{41.18} & \textbf{21.61} & 6.61 & \textbf{23.47} & 48.51 & 50.36 & \textbf{89.73} & \textbf{87.86} & 80.18 & 77.03 & 78.59 \\
+ Mean-shift & 16.76 & 33.28 & 18.81 & 5.16 & 18.50 & 40.89 & \textbf{53.15} & \textbf{89.73} & 87.02 & \textbf{83.24} & \textbf{78.29} & \textbf{79.05} \\
\bottomrule
\end{tabular}
\caption{Llama-3.1-8B on KLAR. Open-ended values are Spearman rank correlation ($\times 100$) with English; multiple-choice values are answer agreement (\%) with English. Acc is factual accuracy (\%). Bold indicates best result per column.}
\label{tab:llama_klar}
\end{table*}

\begin{table*}[t]
\centering
\small
\begin{tabular}{lcccccc|cccccc}
\toprule
& \multicolumn{6}{c|}{Open-ended QA} & \multicolumn{6}{c}{Multiple-choice QA} \\
\cmidrule(lr){2-7} \cmidrule(lr){8-13}
Method & ar-en & en-nl & en-ru & en-zh & Avg & Acc & ar-en & en-nl & en-ru & en-zh & Avg & Acc \\
\midrule
Raw & 14.60 & 26.93 & 3.97 & -0.92 & 11.15 & \textbf{37.89} & 68.61 & \textbf{94.50} & 92.21 & 91.22 & 86.64 & 87.51 \\
PCA & 16.09 & 27.46 & 4.29 & 0.89 & 12.18 & 37.33 & 69.07 & \textbf{94.46} & \textbf{92.36} & 91.33 & 86.81 & 87.60 \\
Mean-shift & 1.92 & 20.58 & 7.33 & 12.95 & 10.70 & 27.50 & \textbf{72.70} & 94.35 & 91.79 & 90.68 & \textbf{87.38} & \textbf{88.25} \\
AE & 15.31 & \textbf{29.90} & 4.98 & -0.84 & 12.34 & \textbf{37.55} & 69.03 & 93.97 & 92.33 & \textbf{91.56} & 86.72 & 87.45 \\
+ PCA & 14.38 & 29.27 & 4.65 & 0.52 & 12.21 & 37.47 & 68.92 & 94.04 & \textbf{92.36} & 91.49 & 86.70 & 87.44 \\
+ Mean-shift & \textbf{18.47} & 18.04 & \textbf{18.21} & \textbf{13.16} & \textbf{16.97} & 30.05 & 68.38 & 93.66 & 91.79 & 91.29 & 86.28 & 87.02 \\
\bottomrule
\end{tabular}
\caption{Qwen-3-8B on KLAR. Open-ended values are Spearman rank correlation ($\times 100$) with English; multiple-choice values are answer agreement (\%) with English. Acc is factual accuracy (\%). Bold indicates best result per column.}
\label{tab:qwen_klar}
\end{table*}

\begin{table*}[t]
\centering
\small
\begin{tabular}{lccccccc|cccccc}
\toprule
& & \multicolumn{6}{c|}{Open-ended QA} & \multicolumn{6}{c}{Multiple-choice QA} \\
\cmidrule(lr){3-8} \cmidrule(lr){9-14}
 & Layer & ar-en & en-nl & en-ru & en-zh & Avg & Acc & ar-en & en-nl & en-ru & en-zh & Avg & Acc \\
\midrule
\multicolumn{2}{l}{Raw} & 21.70 & 38.55 & 19.46 & 6.84 & 21.64 & 45.85 & 48.76 & 89.58 & 87.06 & 79.84 & 76.31 & 78.21 \\
\midrule
\multirow{4}{*}{\rotatebox{90}{Two-phase}}
& L20 & 18.71 & 38.29 & \textbf{20.96} & \textbf{8.84} & \textbf{21.70} & 49.00 & 50.82 & 88.35 & 86.75 & 81.37 & 76.82 & 78.12 \\
& L24 & 21.56 & 39.29 & 19.03 & 5.78 & 21.42 & 48.89 & 50.13 & \textbf{89.77} & \textbf{87.74} & 80.41 & \textbf{77.01} & \textbf{78.55} \\
& L28 & \textbf{22.16} & 37.53 & 17.97 & 6.14 & 20.95 & \textbf{49.14} & 49.37 & 89.00 & 87.13 & 81.02 & 76.63 & 77.99 \\
& L31 & 17.92 & 34.55 & 17.51 & 7.76 & 19.44 & 40.75 & 43.26 & 85.76 & 80.64 & 73.77 & 70.86 & 71.35 \\
\midrule
\multirow{4}{*}{\rotatebox{90}{Joint}}
& L20 & 5.03 & 19.54 & 8.67 & 2.57 & 8.95 & 14.56 & \textbf{61.05} & 82.74 & 50.29 & 51.93 & 61.50 & 49.82 \\
& L24 & 4.24 & 17.13 & 9.37 & 5.37 & 9.03 & 19.34 & 50.25 & 87.63 & 83.66 & 74.84 & 74.10 & 76.09 \\
& L28 & 5.83 & \textbf{40.77} & 14.34 & 7.00 & 16.99 & 35.45 & 49.60 & 86.87 & 86.25 & \textbf{82.21} & 76.23 & 77.57 \\
& L31 & 19.97 & 32.77 & 18.06 & 6.37 & 19.29 & 45.17 & 46.28 & 87.71 & 83.12 & 75.56 & 73.17 & 74.93 \\
\bottomrule
\end{tabular}
\caption{Two-phase versus joint training of the autoencoder and language-specific decoders, evaluated on KLAR with Llama-3.1-8B. Open-ended values are Spearman rank correlation ($\times 100$) with English; multiple-choice values are answer agreement (\%) with English.}
\label{tab:training_ablation}
\end{table*}

\section{Results on Additional Models}
\label{sec:other_models}

To test whether our findings depend on the choice of model, we repeat the KLAR evaluation on two further 8B-parameter models: Llama-3.1-8B, whose pretraining is predominantly English-centric, and Qwen-3-8B. Tables~\ref{tab:llama_klar}--\ref{tab:qwen_klar} report open-ended and multiple-choice results for both.

The main patterns hold across all three models. AE-based intervention improves average consistency with English in both formats, and the most effective intervention layers fall consistently in the final third of the network. The magnitude of the gains varies with how well aligned a model already is: on Llama-3.1-8B, where baseline cross-lingual agreement is lowest, AE+PCA improves open-ended consistency from 21.64 to 23.47, and AE+Mean-shift improves multiple-choice agreement from 76.31 to 78.29, in both cases without a drop in accuracy. On Qwen-3-8B, whose baseline agreement is already high, gains are correspondingly smaller.

\section{Two-Phase vs.\ Joint Training}
\label{sec:training_ablation}

Our AE is trained in two phases: a shared encoder and decoder are optimized jointly, after which the encoder is frozen, and language-specific decoders are trained independently. To verify that this split is necessary, we compare it against training the shared encoder and the language-specific decoders end-to-end in a single phase. Table~\ref{tab:training_ablation} reports KLAR results for both variants on Llama-3.1-8B across four layers.

Two-phase training is clearly the stronger design. Joint training degrades both consistency and accuracy at every layer, and the degradation is severe at earlier layers: multiple-choice accuracy falls from 78.12 to 49.82 at L20, and open-ended accuracy from 49.00 to 14.56. We attribute this to the joint objective, which allows the language-specific decoders to absorb alignment pressure that should instead be resolved in the shared latent space, thereby destabilizing the encoder. Freezing the encoder after the first phase preserves the cross-lingual geometry while still allowing each decoder to specialize.

\end{document}